\documentclass[runningheads]{llncs}
\usepackage{graphicx}
\usepackage{booktabs}
\usepackage{comment}
\graphicspath{{figures/}}
\usepackage{array}

\newcolumntype{P}[1]{>{\centering\arraybackslash}p{#1}}

\begin{document}

\title{Weapon Engagement Zone Maximum Launch Range Estimation Using a Deep Neural Network}

\author{
Joao~P.~A.~Dantas\inst{1}\orcidID{0000-0003-0300-8027} \and  
Andre~N.~Costa\inst{1}\orcidID{0000-0002-2309-9248}\and 
Diego~Geraldo\inst{1}\orcidID{0000-0003-1389-9142} \and 
Marcos~R.~O.~A.~Maximo\inst{2}\orcidID{0000-0003-2944-4476} \and 
Takashi~Yoneyama\inst{3}\orcidID{0000-0001-5375-1076}}

\authorrunning{J. P. A. Dantas et al.}
\institute{Decision Support Systems Subdivision, Institute for Advanced Studies, \\
Sao Jose dos Campos - SP 12.288-001, Brazil \\
\email{\{dantasjpad,negraoanc,diegodg\}@fab.mil.br}\\
\url{http://www.ieav.cta.br} \and
Autonomous Computational System Lab (LAB-SCA), Computer Science Division, Aeronautics Institute of Technology, Sao Jose dos Campos - SP 12228-900, Brazil \\
\email{mmaximo@ita.br} \\
\url{http://www.comp.ita.br/labsca/} \and
Electronic Engineering Division, Aeronautics Institute of Technology, \\
Sao Jose dos Campos - SP 12228-900, Brazil \\
\email{takashi@ita.br}\\
\url{http://www.ele.ita.br}}

\titlerunning{WEZ Maximum Launch Range Estimation using a DNN}

\maketitle

\begin{abstract}
This work investigates the use of a Deep Neural Network (DNN) to perform an estimation of the Weapon Engagement Zone (WEZ) maximum launch range. The WEZ allows the pilot to identify an airspace in which the available missile has a more significant probability of successfully engaging a particular target, i.e., a hypothetical area surrounding an aircraft in which an adversary is vulnerable to a shot. We propose an approach to determine the WEZ of a given missile using 50,000 simulated launches in variate conditions. These simulations are used to train a DNN that can predict the WEZ when the aircraft finds itself on different firing conditions, with a coefficient of determination of 0.99. It provides another procedure concerning preceding research since it employs a non-discretized model, i.e., it considers all directions of the WEZ at once, which has not been done previously. Additionally, the proposed method uses an experimental design that allows for fewer simulation runs, providing faster model training.

\end{abstract}

\section{Introduction}
Within simulated computational environments, military systems must resemble reality in a level of fidelity that leads to useful conclusions~\cite{hancock2008human}. This is done through the use of reliable computational models, that are deemed to encompass the main characteristics of the systems they represent~\cite{hill2001applications}.

When dealing with air combat, one of the most critical parts to be modeled is the missile. This is true concerning both the missile system itself and the decision of when to employ it, i.e., to fire. That is even more critical when considering Beyond Visual Range (BVR) air combat since this decision must be taken based only on what the situational awareness systems display to the pilot~\cite{dantas2018}.

In the context of constructive simulations, in which the aircraft behave autonomously, there is a need to provide their controlling algorithms with data similar to what real pilots would receive, so that the behaviors perform in accordance~\cite{costa2019master}. One of the most important aspects that a pilot can use to decide whether to launch a missile on an opposing aircraft is the Weapon Engagement Zone (WEZ), which, in simple terms, represents the range of the weapon~\cite{dantaslars2021}. This definition is discussed with more depth further in Section~\ref{wez}. The determination of this range is not a simple task, however, since it is influenced by a series of variables from both the shooter and the target. Moreover, it is naturally dependent on the missile itself. In this work, we propose an approach to determine the WEZ of a given missile using a series of simulated launches in variate conditions. These simulations are used to train a machine learning algorithm that can predict the WEZ when the aircraft finds itself on different firing conditions. Previous works have employed some types of Artificial Neural Networks (ANN), such as Wavelet Neural Networks (WNN)~\cite{yoon2010new} and a Multi Layer Perceptron (MLP) with Bayesian Regularization of Artificial Neural Networks (BRANN)~\cite{birkmire2012air}, to make predictions of the WEZ, also from previously simulated data. Purely mathematical approaches are also available within the literature, such as~\cite{farlik2017simplification} and~\cite{li2020simulation}, but they provide an intermediate step between unrealistic missile models that consider fixed missile ranges and more complex models based on simulations.

Much more research may have been developed within companies and governments concerning WEZ determination~\cite{birkmire2011weapon}, but this is still seldom publicly available. The contribution of this work is employing a Deep Neural Network (DNN) with a novel non-discretized model, i.e. the model considers all directions of the WEZ at once, not discretizing the off-boresight angle (Fig.~\ref{fig5}) as done previously to the best of our knowledge. Additionally, it uses an experimental design that allows for a lower number of simulation runs, which provides a faster training of the model.

The remainder of this paper is organized as follows. Section~\ref{Background} provides the background, explaining in more depth the concept of WEZ, as well as presenting the particular missile model employed and the experimental design utilized. In Section~\ref{Method}, the proposed methodology is detailed, whereas the results coming from it are presented and analyzed in Section~\ref{Results}. Finally, Section~\ref{Conclusion} states the main conclusions of the work and suggests some future developments.

\section{Background}\label{Background}

In this section, we detail the concept of WEZ, present the missile model, and specify the simulation experimental design used within this work.

\subsection{Weapon Engagement Zone}\label{wez}
The term WEZ may present different definitions throughout the military domain. According to the United States Department of Defense~\cite{dod2021}, WEZ can be described as an “airspace of defined dimensions within which the responsibility for engagement of air threats normally rests with a particular weapon system.” Although being a rather broad definition, its focus resides on the responsibility for engagement of target that is inside the zone by a specific system.

In our work, on the other hand, we are more focused on the airspace defined by the range of a weapon system (missile), which is not necessarily responsible for engaging all threats within this zone. This is rather a possibility, that is, the WEZ in our case allows the pilot to identify an airspace in which the missile available has a larger probability of being successful in engaging a particular target. In other words, the definition of WEZ adopted by us is similar to what Portrey \emph{et al.}~\cite{portrey2005pairwise} present: a hypothetical area surrounding an aircraft in which an adversary is vulnerable to a shot. This concept can be found in the literature under different terminologies which may present subtle variations on meaning, such as Launch Acceptability Region (LAR)~\cite{yoon2010new} and Dynamic Launch Zone (DLZ)~\cite{alkaher2015dynamic}.

\begin{figure}
\centering
\includegraphics[width=0.7\textwidth]{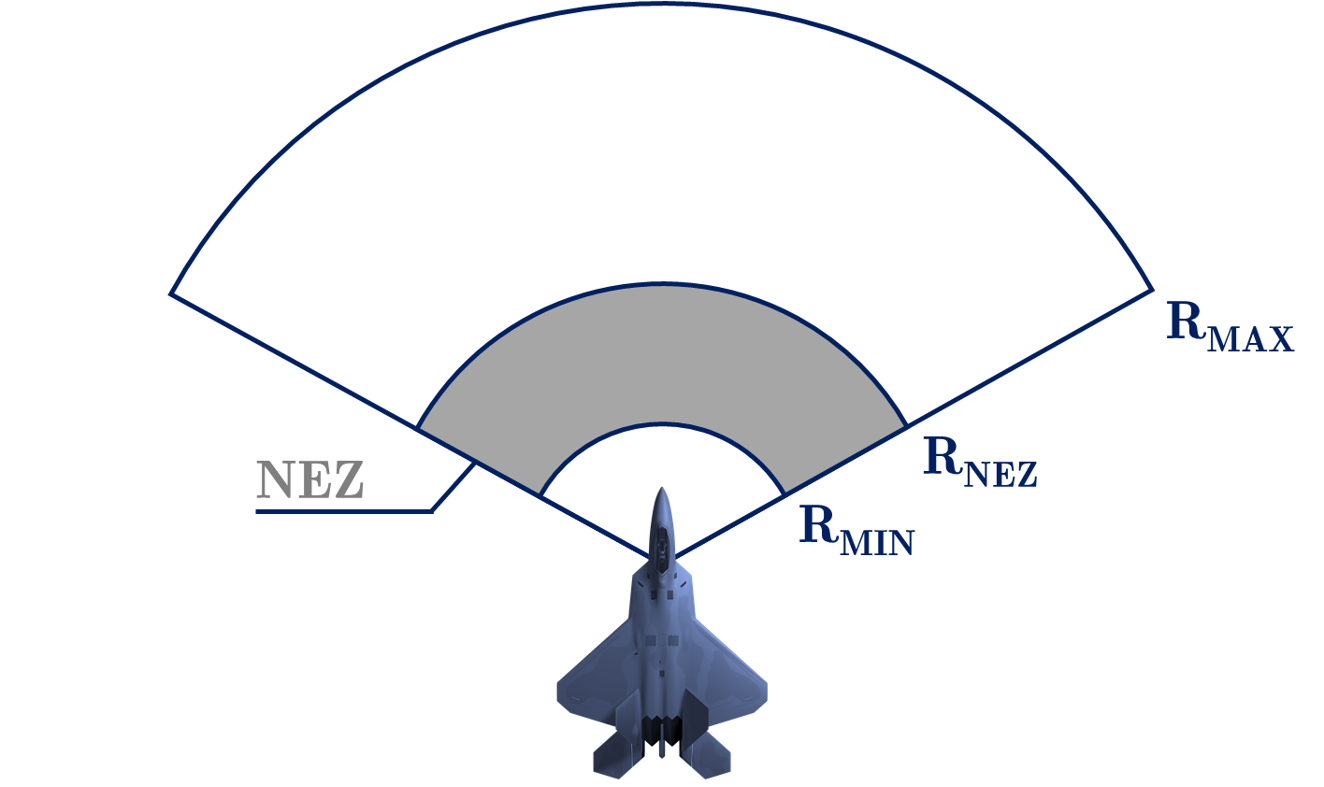}
\caption{Simplified WEZ representation.} \label{fig1}
\end{figure}

Fig.~\ref{fig1} presents a simplified depiction of a WEZ, which stretches from the minimum range \(R_{min}\) to the maximum range \(R_{max}\). The \(R_{max}\) is defined by us as the maximum distance in which the missile will hit a non-maneuvering target, that is, if the target performs any maneuver, the missile will most likely miss if fired at this distance. On the other limit of this zone, the \(R_{min}\) is the minimum distance required by the missile to be able to properly activate its systems and, therefore, trigger its warhead. Between these two ranges, there is the no-escape zone (NEZ) range (\(R_{NEZ}\)), which represents a distance within which the target is very unlikely to be able to evade the missile, even when employing a high-performance defensive maneuver.

It is important to point out that the WEZ is also a function of the threat since it takes into consideration the parameters of the target in its calculation. As Portrey \emph{et al.}~\cite{portrey2005pairwise} state, the WEZ is determined by many factors regarding both the shooter and the target, such as “type of weapon, aircraft speed, relative altitudes, and geometry.”  These factors are used by the authors of~\cite{portrey2005pairwise} to define a metric that allows the pilot to know what is the amount of G-force that must be pulled to escape from an incoming missile. Therefore, their focus was less on the definition of the WEZ per se, but rather on the determination of this particular metric.

\begin{figure}
\centering
\includegraphics[width=\textwidth]{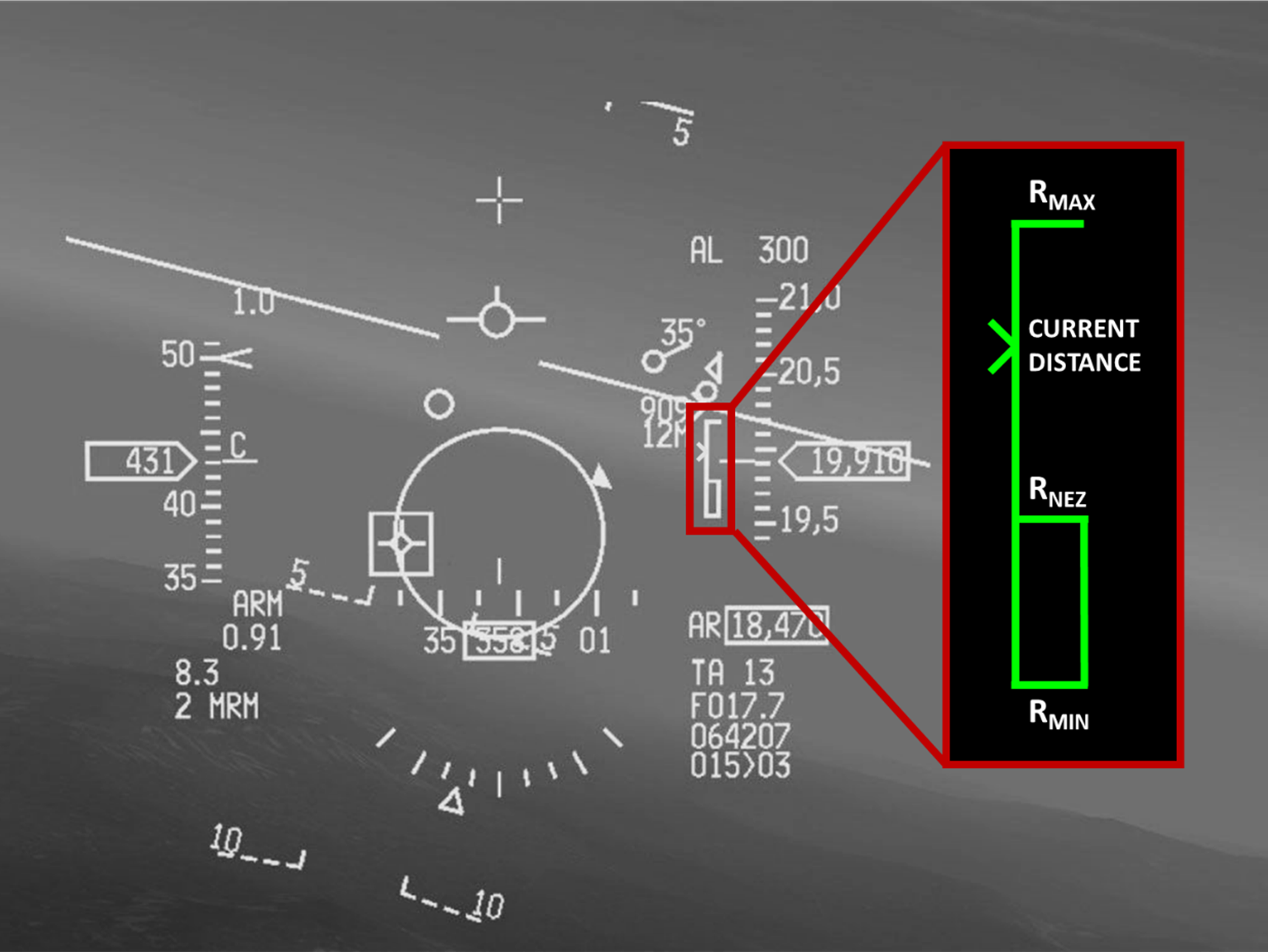}
\caption{HUD representation with focus on the WEZ indication.} \label{fig2}
\makebox[\width]{Source: Adapted from \cite{sitehud}}.
\end{figure}

On the other hand, Birkmire~\cite{birkmire2011weapon}, focuses precisely on the determination of the WEZ for a missile in the context of virtual simulations, i.e., simulations in which real pilots interact with simulated systems. Therefore, his goal was to provide the pilots in virtual environments with a similar estimation of the WEZ as pilots in real aircraft have in their heads-up displays (HUDs) to support their decisions to fire a missile (Fig.~\ref{fig2}).

Our work has a slightly different focus since it aims to provide WEZ information to autonomous agents within a constructive simulation environment, i.e., a simulation in which simulated pilots interact with simulated systems. In addition, we provide a map visualization of the estimated WEZ, which can be valuable within the analysis of this type of simulation.

\subsection{Missile Model}
\begin{figure}
\includegraphics[width=\textwidth]{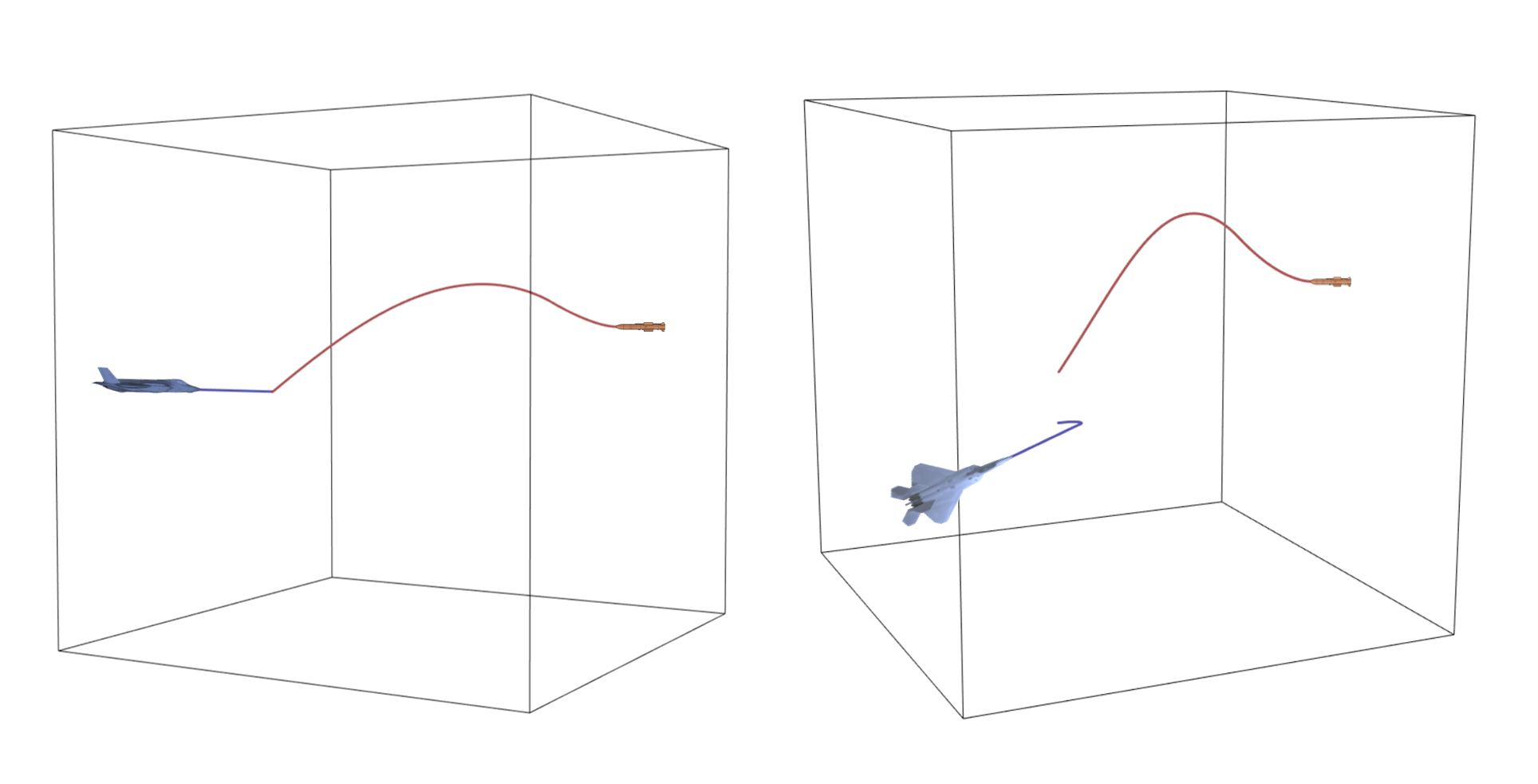}
\caption{Missile simulated trajectory samples.} \label{fig3}
\end{figure}

Since this is not the focus of this work, inasmuch as the methodology presented may be applied to any simulated missile, we just provide a brief overview of the missile model. Our implementation is completely done in the R programming language~\cite{ihaka1996r} and it provides a simplified model with 5 degrees of freedom (5DOF) of a Fox 3 missile-based on~\cite{handbook1995missile}. According to~\cite{brevity2020}, Fox is a brevity code that refers to the guidance of a missile, in which type 3 stands for an active radar-guided missile, i.e., a missile which contains a seeker of its own that can track the target autonomously after reaching its activation distance. Still, with regards to its guidance, the missile performs perfect proportional navigation concerning its target, maneuvering to exactly comply with its guidance law, as well as a loft maneuver (i.e., an aggressive climb right after launch) whenever possible, as Fig.~\ref{fig3} shows. 

\begin{figure}
\includegraphics[width=\textwidth]{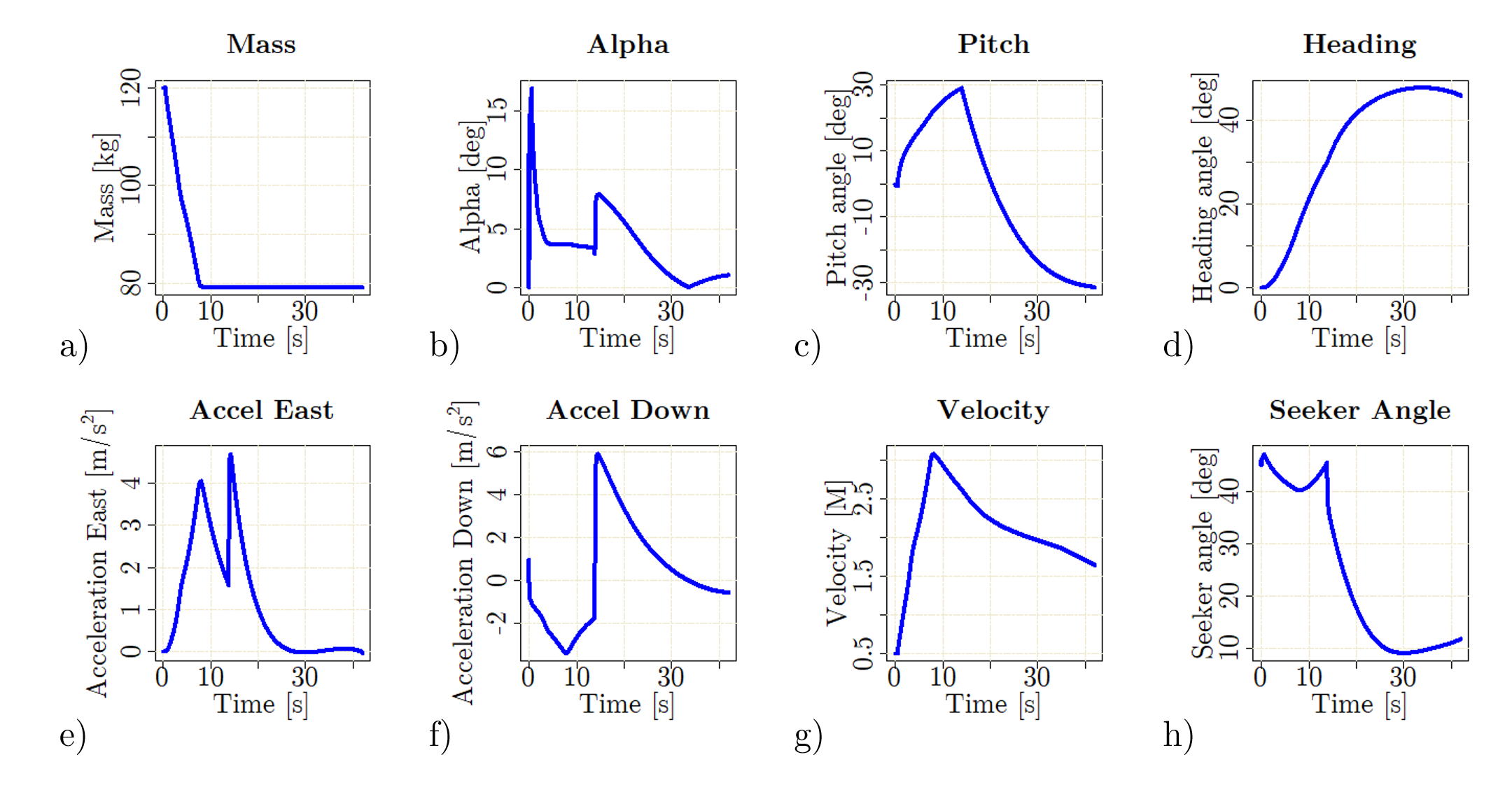}
\caption{Missile trajectory metrics samples.} \label{fig4}
\end{figure}

The model simulates the missile trajectory considering either a still or a maneuvering target. To define the NEZ range, the simulation considers a high-performance maneuver of $+5$ G, which may be employed with a delay from the moment of launch. Some important metrics for the missile flight are provided in Fig.~\ref{fig4}.

Referring to Fig.~\ref{fig4}, the most straightforward metric is the mass (a). Since the missile operates with a boost-sustain motor~\cite{noaman2020boost}, its mass decays almost linearly during its boost (burn) phase. Due to the loft maneuver, angle of attack values (b) vary very aggressively at the beginning of the flight, which can also be observed on the pitch angle (theta) chart (c). Concerning heading (psi), there are some maneuvers to respond to the high-performance evasion that the target employs (d). Accelerations in the East (e) and Down (f) axis in the NED coordinate system~\cite{cai2011coordinate} are also very abrupt due to the loft maneuver and the target response, respectively.  The velocity (g) steadily increases during burn time and decays on the sustain phase. Finally, the seeker angle (h) accounts for the proportional navigation, being defined as the deviation of the shooter’s longitudinal axis from the off-boresight angle (Fig.~\ref{fig5}).

\subsection{Experimental Design}
\begin{figure}
\centering
\includegraphics[width=0.965\textwidth]{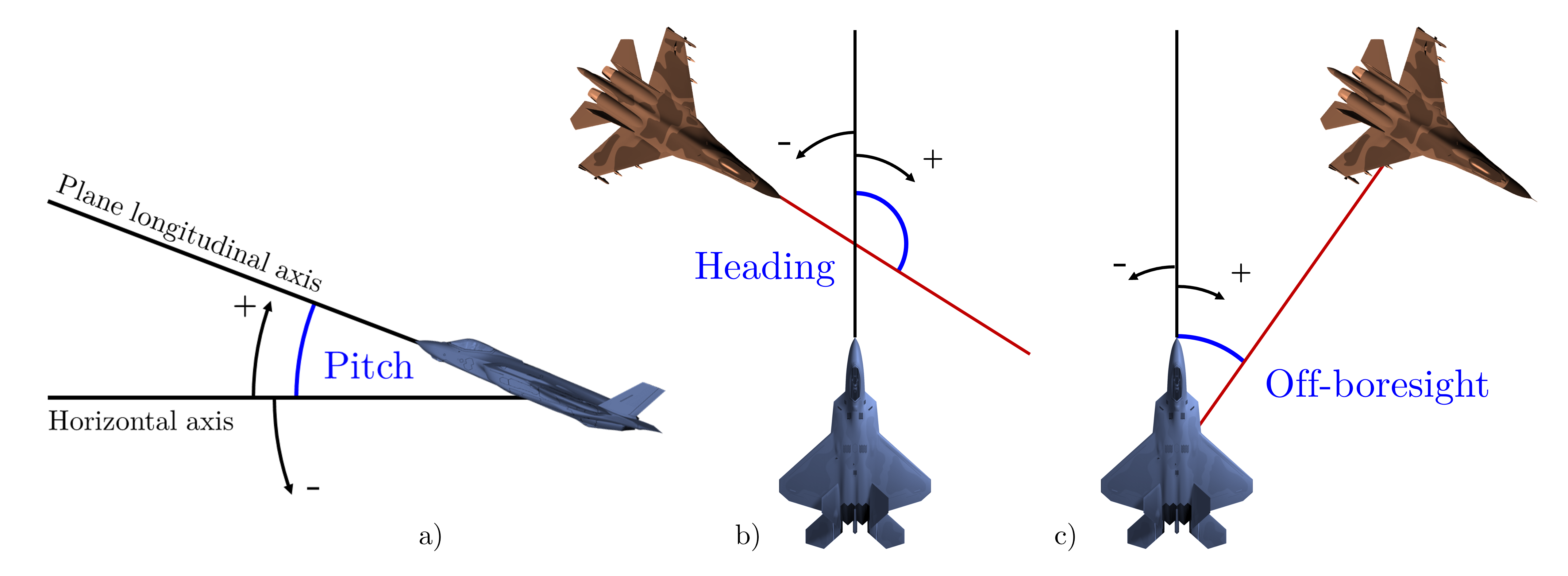}
\caption{Pitch (a), heading (b), and off-boresight (c) angles with respect to the target aircraft.} \label{fig5}
\end{figure}

The parameters used as inputs to our missile model (Table~\ref{tb1}) are very similar to the ones presented in~\cite{birkmire2011weapon}, which makes it easier to compare our results with the ones obtained by it. However, instead of using an implementation in MATLAB Simulink~\cite{klee2018simulation}, our model was implemented entirely on R language as aforementioned, which has many prepackaged programs that help to solve analytical problems, prioritizing the simplicity of understanding and the parametrization. To provide a common understanding of the angles used, Fig.~\ref{fig5} provides a depiction of them.

These parameters are selected based on operational experience and the missile model possibilities. The shooter's velocity and altitude are directly related to the energy that will be available to the missile. In particular, the launch altitude also influences the drag to which the missile will be subjected during flight, which is also true concerning the target altitude on the missile final approach. Target's velocity can either help or hinder the missile's effectiveness, depending on its heading. However, heading alone cannot provide a full account with regards to positioning, since this is dependent on the off-boresight angle to determine whether the target aircraft is getting closer to the shooter and, therefore, to the missile itself. At last, the shooter's pitch angle at the moment of launch may help the initial maneuvering of the missile, that is, its loft maneuver.

Instead of a full factorial experiment as~\cite{birkmire2011weapon} presented, we tried to reduce the number of simulation runs by means of a more sophisticated design that takes into account randomness in its formation. Alternatively of a Monte Carlo simulation (MCS) that simply randomly samples the search space~\cite{homem2014monte}, we used the Latin Hypercube Sampling (LHS), which is deemed to be more efficient~\cite{deutsch2012latin}. 

The LHS is a near-random method, which aims at a better coverage of the search space, since a purely random approach may concentrate the samples by chance. Its main idea is to divide the multidimensional space so that the random samples are drawn from these subdivisions instead of the whole search space~\cite{husslage2011space}. In our particular case, we employed a maximin algorithm, which attempts to optimize the sample through the maximization of the minimum distance between design points, fulfilling the constraints established by the LHS method.
Table~\ref{tb1} presents the intervals for each variable used in the sampling. These limits were defined by subject matter experts, in this case, pilots, which considered meaningful values concerning their operational context.

\begin{table}[h]
\centering
\caption{Model parameters with the respective intervals considered.}\label{tb1}
\begin{tabular}{|l|c|c|c|c|}
\hline
\textbf{Parameter} & \textbf{Variable} & \textbf{Min} & \textbf{Max} & \textbf{Unit} \\
\hline
Shooter altitude     & alt\_sht & 1,000 & 45,000 & feet    \\
Shooter velocity     & vel\_sht & 400  & 600   & knots   \\
Shooter pitch        & pit\_sht & -45  & 45    & degrees \\
Target altitude      & alt\_tgt & 1,000 & 45,000 & feet    \\
Target velocity      & vel\_tgt & 400  & 600   & knots   \\
Target heading       & hdg\_tgt & -180 & 180   & degrees \\
Target off-boresight & rgt\_tgt & -60  & 60    & degrees \\
\hline
\end{tabular}
\end{table}

\section{Methodology}\label{Method}

This section contains the description of the preprocessing, training and evaluation of the DNN model that is applied on the data coming from the simulation.

\subsection{Simulation}
After creating the input batch files, through LHS and with the limits presented previously, 50,000 simulations were run using 2 Intel Xeon Silver 4210R CPUs with 2.40GHz and 128 GB of RAM. It took approximately 7 hours to execute all the simulations, which generated an output file containing the maximum range of the missile for the respective input conditions.

\subsection{Preprocessing}
From that, an Exploratory Data Analysis (EDA) was performed to identify general behaviors of the output data. The methods employed in this analysis were: histogram,  boxplots, correlation, and descriptive statistics.

Before performing the training of the ANN, some feature engineering techniques were employed. The first one was a form of encoding to better deal with cyclical features. The angles related to aircraft heading and off-boresight were encoded into their sine and cosine counterparts as done in~\cite{petnehazi2019recurrent}, slightly increasing our model performance.

In addition, a form of handling potential outliers was to perform downsampling of the Latin Hypercube design. This was done because the pre-established intervals generated some improbable conditions. For instance, an aircraft at 1,000 ft firing on a target at 45,000 ft is exceedingly rare from the operational standpoint since a pilot would most likely increase its altitude before launching a missile. Therefore, we removed these undesirable samples, like the one presented, from the whole dataset based on subject matter expert operational knowledge, which can vary according to the mission type.

Lastly, data scaling was performed to equally distribute the importance of each input in the ANN learning process~\cite{priddy2005artificial}. This was done through a min-max scaler, which individually scales and translates all data features to a range from 0 to 1~\cite{bonaccorso2017machine}.

\subsection{Model training}
\begin{figure}
\includegraphics[width=\textwidth]{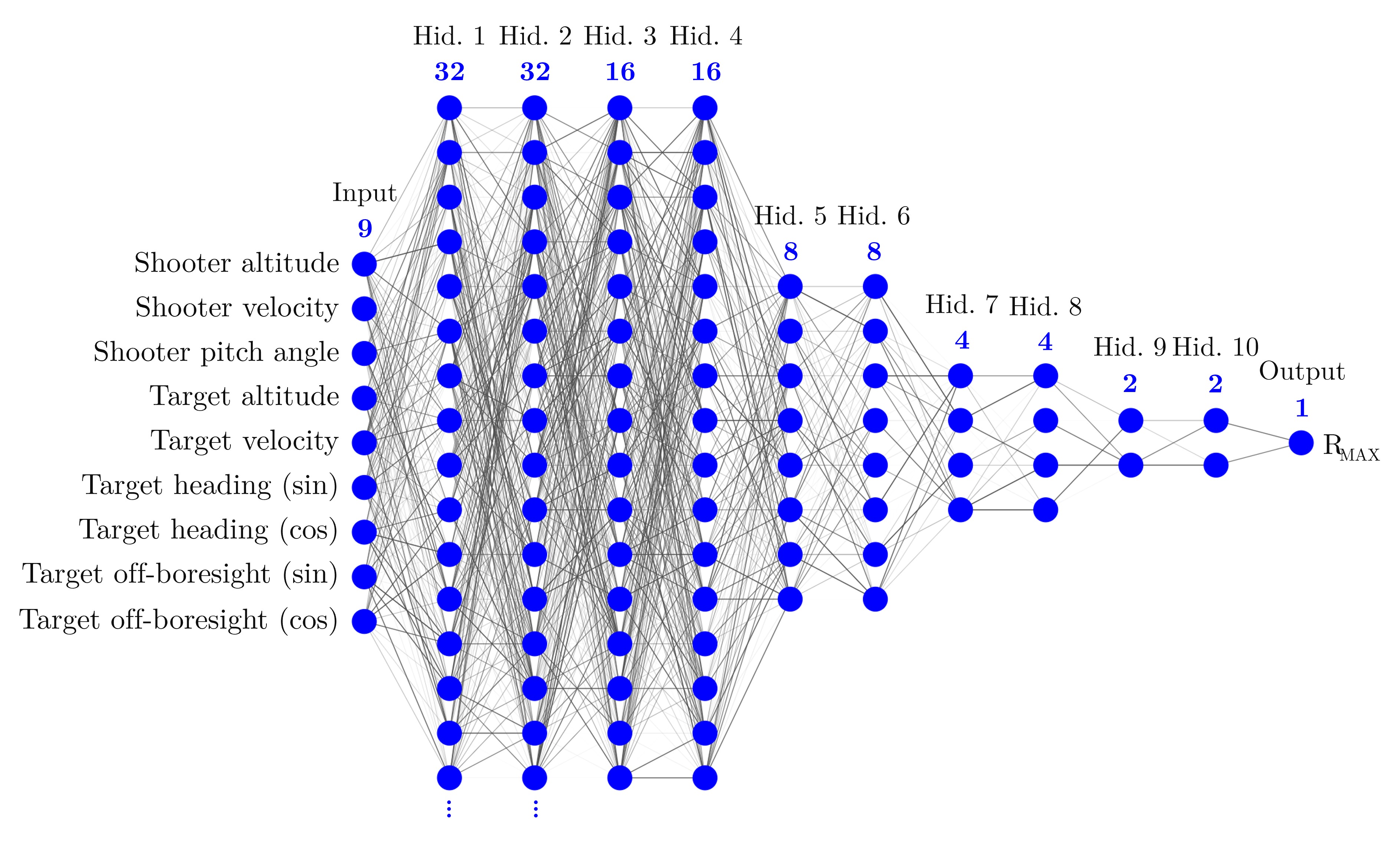}
\caption{Proposed DNN architecture.} \label{fig10}
\end{figure}

Before training the DNN, a train-validate-test split was performed, allocating 80\% for training and validation using a 5-fold cross-validation technique, and 20\% for testing. This division is done randomly and will allow the evaluation of the machine learning model later. The DNN model was formed by 12 layers of nodes, with the structure represented in Fig.~\ref{fig10}. All nodes have a rectified linear activation function (ReLU)~\cite{bengio2017deep}.

In addition, the Adaptive Moment Estimation (Adam) optimizer was employed, an extremely popular training algorithm for ANN~\cite{bock2019proof}. Adam is a stochastic gradient descent method based on adaptive estimation of first- and second-order moments function~\cite{kingma2014adam}, which, in our case, aimed to minimize the Mean-Squared Error (MSE) loss. This was monitored by an early stopping method that checked whether the validation set metric had stopped improving (the patience, i.e., the number of epochs to wait before early stop if no progress on the validation set, was set to 20).

\subsection{Model evaluation}

As the model being analyzed deals with a regression problem, the evaluation of the model will be carried out observing the following metrics: Mean Absolute Error (MAE), Mean Squared Error (MSE), Root Mean Squared Error (RMSE), and coefficient of determination ($R^{2}$).

\section{Results and analysis}\label{Results}

This section examines the exploratory data analysis and the test dataset metrics. Additionally, it provides a Multi-Function Display (MFD) representation, focusing on the WEZ indication based on the proposed model.

\subsection{Exploratory Data Analysis}

\begin{figure}
\centering
\includegraphics[width=\textwidth]{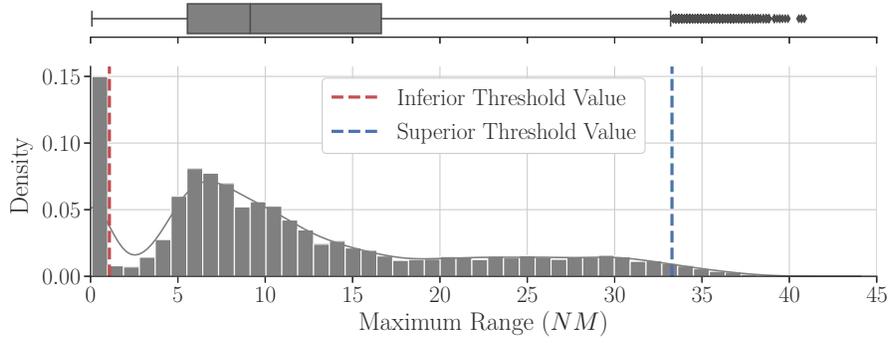}
\caption{Histogram and boxplot of the target variable.} \label{fig8}
\end{figure}

Initially, an overview of the descriptive statistics of the model's input and output variables was observed, as shown in Table \ref{tb2}. The input variables of the model follow a uniform distribution since these variables were sampled using the LHS. The model's output variable presents great variability with an average of 12.38 NM and a standard deviation (std) of 9.37 NM. Notice that the mean and median (50\%) are varying by 3.24 NM, which indicates a considerable amount of outliers for this variable at the top of the distribution. These outliers will be eliminated from a superior threshold value (33.28 NM), which is not the maximum value (max), but is rather the largest value of the sampling excluding outliers, based on the interquartile range ($75\% - 25\%$). Observing the minimum (min), values of the order of 0.08 NM can be found, which shows that in the dataset there are values in the target variable (\texttt{max\_range}) that are smaller than the minimum activation distance of the missile modeled. For this case, this distance is considered to be 2 km ($1.079$ NM), which is the inferior threshold. So that the model would not be harmed in its training to try to predict the maximum missile range distance values, samples in which the model's output variable was smaller than the minimum missile activation distance were removed from the dataset. A histogram and a boxplot were generated together to visualize the distribution and the thresholds of the target variable, which can be seen in Fig.~\ref{fig8}.

\begin{table}
\centering
\caption{Descriptive statistics of the model's input and output variables.}\label{tb2}
\begin{tabular}{|c|c|c|c|c|c|c|c|c|}
\cline{2-9}
\multicolumn{1}{c|}{} &
 \textbf{\begin{tabular}[c]{@{}c@{}}alt\_sht\\ (ft)\end{tabular}} &
  \textbf{\begin{tabular}[c]{@{}c@{}}vel\_sht\\ (kt)\end{tabular}} &
  \textbf{\begin{tabular}[c]{@{}c@{}}pit\_sht\\ (deg)\end{tabular}} &
  \textbf{\begin{tabular}[c]{@{}c@{}}alt\_tgt\\ (ft)\end{tabular}} &
  \textbf{\begin{tabular}[c]{@{}c@{}}vel\_tgt\\ (kt)\end{tabular}} &
  \textbf{\begin{tabular}[c]{@{}c@{}}hdg\_tgt\\ (deg)\end{tabular}} &
  \textbf{\begin{tabular}[c]{@{}c@{}}rgt\_tgt\\ (deg)\end{tabular}} &
  \textbf{\begin{tabular}[c]{@{}c@{}}max\_range\\ (NM)\end{tabular}} \\ 
 \hline
\textbf{mean}  & 23,000.00 & 500.00   & 0.00    & 23,000.00 & 500.00   & 0.00     & 0.00     & 12.38    \\
\textbf{std}   & 12,701.83 & 57.74	 & 25.98	& 12,701.83	& 57.74	& 103.92 &	34.64 & 	9.37     \\
\textbf{min}   & 1,000.22  &	400.00	 & -45.00	& 1,000.82	& 400.00	& -180.00	& -60.00 & 	0.08 \\
\textbf{25\%}  & 12,000.34 &	450.00	 & -22.50	& 12,000.32	& 450.00	& -90.00	& -30.00	& 5.55     \\
\textbf{50\%}  & 22,999.96 &	500.00	 & 0.00	    & 22,999.99	& 500.00	& 0.00 &	0.00 &	9.14   \\
\textbf{75\%}  & 33,999.75 &	550.00	 & 22.50	& 33,999.76	& 550.00	& 90.00	& 30.00	& 16.64    \\
\textbf{max}   & 44,999.38 &	600.00	 & 45.00	& 44,999.42	& 600.00	& 179.99 & 	60.00 &	40.87    \\
\hline
\end{tabular}
\end{table}

\begin{figure}
\centering
\includegraphics[width=\textwidth]{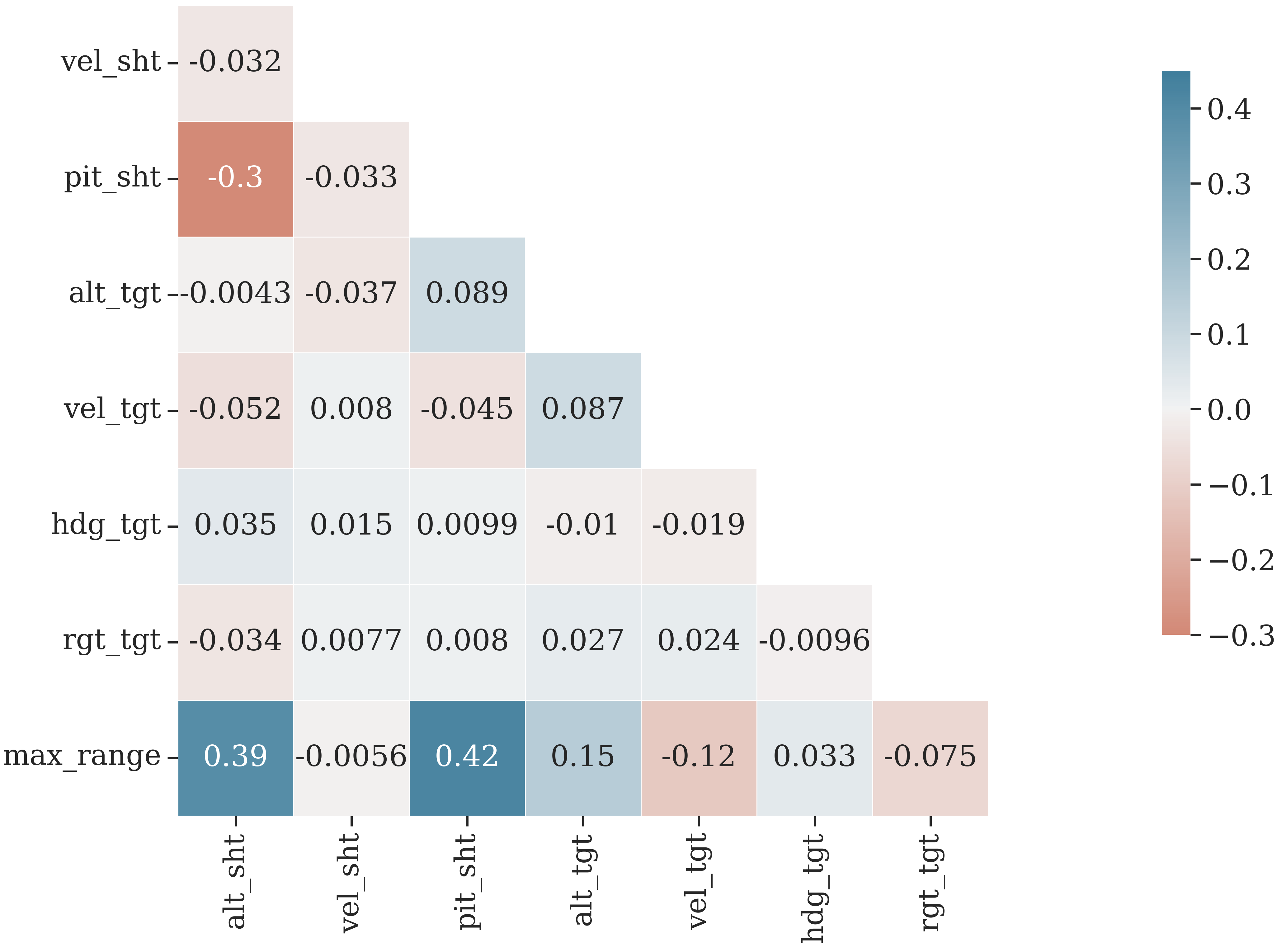}
\caption{Pearson correlation matrix of all model variables.} \label{fig9}
\end{figure}

Pearson's correlation analysis of the variables can be seen in the correlation matrix represented in Fig.~\ref{fig9}. Notice that none of the model's features has a strong correlation with each other, with the largest absolute value being only 0.30 between shooter's altitude (\texttt{alt\_sht}) and pitch (\texttt{pit\_sht}). The performance of the algorithm may deteriorate if two or more variables are tightly related, called multicollinearity. We may also be interested in the correlation between input variables with the output variable (\texttt{max\_range}) to provide insight into which variables may or may not be relevant as input for developing a model. Only the variables \texttt{alt\_sht} and \texttt{pit\_sht} have a slight correlation with the target variable.

\subsection{Model Predictions}
Table~\ref{tb3} shows all the metrics used to evaluate the model with respect to the test set at the end of the training process. Very satisfactory results were found, with a coefficient of determination above $99\%$, which shows a very consistent model. In addition, note that the MAE was around $0.58$ NM, which can be considered a very low value, considering that the values of the target variable have a mean of $13.13$ NM with a standard deviation (std) of $8.58$ NM. If we consider the RMSE, which penalizes the outliers' effects, the observed value is around $1.10$ NM.

\begin{table}
\centering
\caption{Metrics used to evaluate the DNN model at the end of the training process.}\label{tb3}
\begin{tabular}{|P{2.2cm}|P{2.2cm}|P{2.2cm}|P{2.2cm}|}
\hline
\textbf{\begin{tabular}[c]{@{}c@{}}MAE (NM)\end{tabular}} &
\textbf{\begin{tabular}[c]{@{}c@{}}MSE (NM$^{2}$)\end{tabular}} &
\textbf{\begin{tabular}[c]{@{}c@{}}RMSE (NM)\end{tabular}} & \textbf{\begin{tabular}[c]{@{}c@{}}$R^{2}$\end{tabular}} \\
\hline
0.58 &
1.23 &
1.10 &
0.99 \\
\hline
\end{tabular}
\end{table}

A 5-fold cross-validation was conducted to estimate the skill of a machine learning model on unseen data and will help to better understand our data, giving much more information about our algorithm performance. The metrics of the five-folds were very similar as shown in Table~\ref{tb4}. The low variance found between the folds of this sample demonstrates the consistency of the model.

\begin{table}
\centering
\caption{5-fold cross-validations metrics.}\label{tb4}
\begin{tabular}{|P{2.2cm}|P{2.2cm}|P{2.2cm}|P{2.2cm}|P{2.2cm}|}
\cline{2-5}
\multicolumn{1}{c|}{} &
\textbf{\begin{tabular}[c]{@{}c@{}}MAE  (NM)\end{tabular}} &
\textbf{\begin{tabular}[c]{@{}c@{}}MSE  (NM$^{2}$)\end{tabular}} &
\textbf{\begin{tabular}[c]{@{}c@{}}RMSE  (NM)\end{tabular}} & \textbf{\begin{tabular}[c]{@{}c@{}}  $R^{2}$  \end{tabular}} \\
\hline
\textbf{1º Fold}                     & 0.54 & 1.06 & 1.03 & 0.99 \\
\textbf{2º Fold}                     & 0.62 & 1.22 & 1.10 & 0.99 \\
\textbf{3º Fold}                     & 0.71 & 1.39 & 1.18 & 0.98 \\
\textbf{4º Fold}                     & 0.52 & 1.08 & 1.04 & 0.99 \\
\textbf{5º Fold}                     & 0.57 & 1.34 & 1.16 & 0.98 \\
\hline
\textbf{mean} & 0.59 & 1.22 & 1.10 & 0.99 \\
\textbf{std} & 0.08 & 0.15 & 0.07 & 0.01 \\
\hline
\end{tabular}
\end{table}

\subsection{Model Representation}

\begin{figure}
\centering
\includegraphics[width=0.99\textwidth]{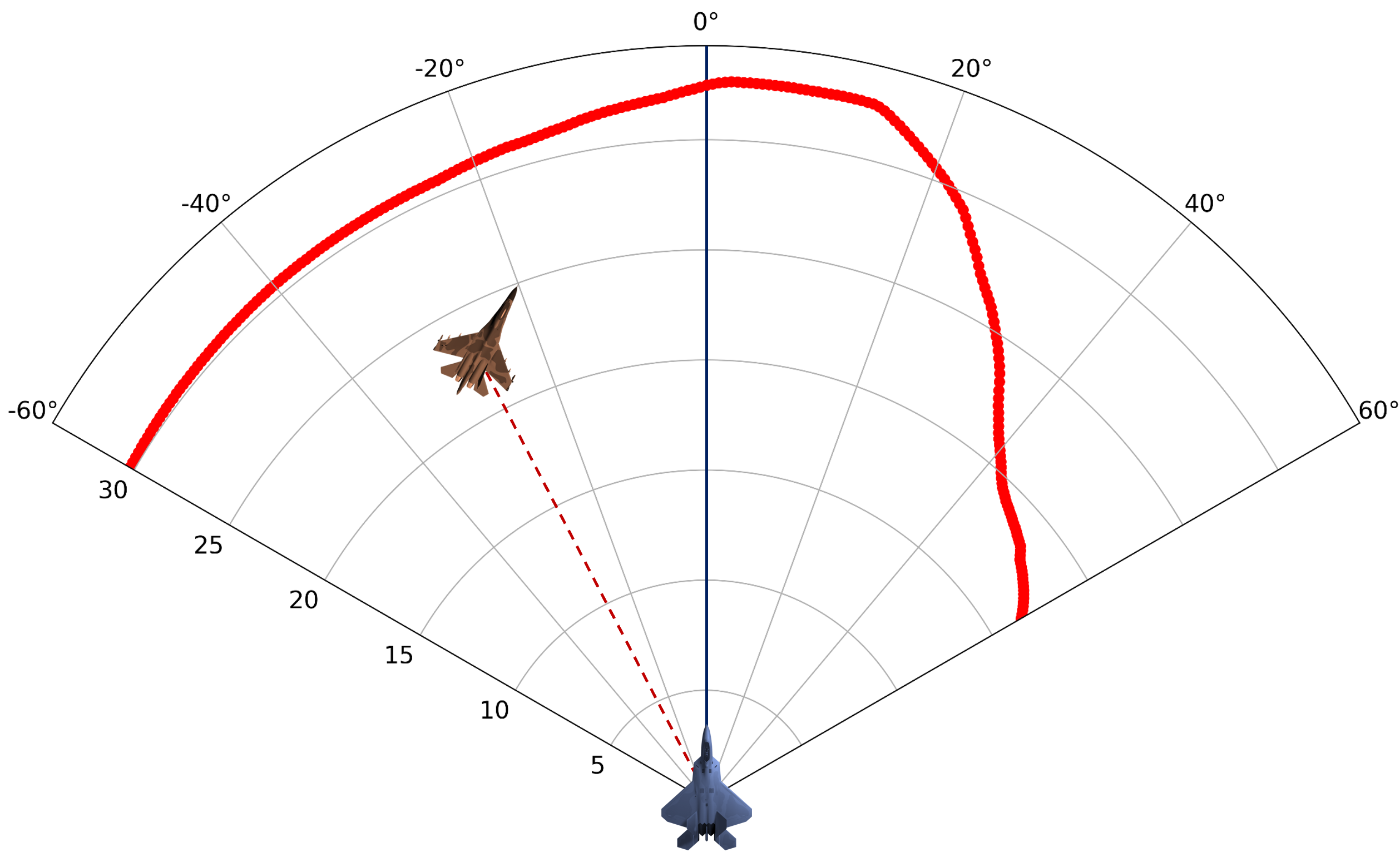}
\caption{MFD representation with a focus on the WEZ maximum range indication.} \label{fig12}
\end{figure}

We estimated the WEZ Maximum Launch Range from the trained model using one of the samples from the test group. The target's position was varied by changing the off-boresight values from $-60^{\circ}$ to $+60^{\circ}$ with steps of $0.5^{\circ}$. A MFD representation with a focus on the WEZ indication can be seen in Fig.~\ref{fig12}. The curve that shows the missile's maximum range proved to be quite consistent, with a continuous aspect throughout the variations of off-boresight angles. Thus, we conclude that employing a different approach, unlike other research, with the incorporation of the off-boresight angle between the reference and the target aircraft as a feature in the model does not affect the performance of the WEZ estimation significantly since the supervised learning model used can be able to generalize well the results obtained in the training dataset to the test dataset.

\section{Conclusions and Future Work}\label{Conclusion}

Therefore, the main improvements advocated by us with respect to~\cite{birkmire2011weapon} is that, instead of discretizing the WEZ calculation concerning the off-boresight angle between shooter and target (Fig.~\ref{fig4}), creating, therefore, several ANNs, we approached the problem considering the whole space defined by the shooter's radar, with only one DNN being able to predict the values of WEZ. In addition, the number of simulation runs was much lower (50,000 runs, as opposed to $222$ million in~\cite{birkmire2011weapon}), which was achieved by a more carefully tailored experimental design.

In addition, in this work, a DNN with an MLP architecture was used, and brought better results than an ANN with only one hidden layer, as done in~\cite{birkmire2011weapon}, comparing the coefficient of determination of both approaches applied to their respective datasets. In addition, different configurations of training and test groups were used in the dataset using the k-fold cross-validation. For a value of $k = 5$, that is, the training and test group was sampled 5 times, the results found were quite similar among all samples, demonstrating the consistency of the DNN model presented in this work.

The use of feature engineering techniques, with the creation of other model input variables, such as the use of sine and cosine for the variables that represent heading and off-boresight angles, also contributed to greater adequacy of the model to the dataset collected from the simulations.
Furthermore, it was observed, with the use of operational knowledge, that some of the samples collected would be unlikely to occur in a real air combat environment. These cases were when, for example, at a given altitude, the speeds of a given agent should meet at certain speed intervals. In the dataset some samples were not respecting these intervals, which could impair the model's performance, trying to predict cases that would most likely not occur in a real situation. To avoid these problems, these samples were eliminated from the dataset.

Future work should investigate how possible improvements in the architecture used for the DNN can bring better results and be more efficient, i.e. with a lower computational cost in the training process. In addition, the results found in this work can be compared with the use of other supervised machine learning techniques. These comparisons will help to determine the most appropriate methodology for calculating WEZ. In addition, in future work, it is possible to carry out calculations not only of the maximum range but also the distances related to the NEZ or even intermediate distances that could provide pilots with more assertive information about the probabilities of a missile reaching its target. Also, more advanced simulation models of the missile may be used in the future to provide better reliability to the presented results.

\section*{Acknowledgments}

This work was supported by Finep (Reference nº 2824/20). Takashi Yoneyama is partially funded by CNPq -- National Research Council of Brazil through the grant 304134/2-18-0. 

\bibliographystyle{splncs04.bst}
\bibliography{wezreferences}

\end{document}